\pdfoutput=1

\documentclass[11pt]{article}

\usepackage[]{acl}

\usepackage{times}
\usepackage{latexsym}
\usepackage{multirow}
\usepackage{comment}
\usepackage{paralist}
\usepackage{enumitem}
\usepackage{etaremune}
\usepackage{color,framed}
\usepackage{xcolor,colortbl}
\usepackage[T1]{fontenc}

\usepackage[utf8]{inputenc}
\usepackage{txfonts}

\usepackage{microtype}

\usepackage{graphicx}
\graphicspath{{graphics/}}
\usepackage{xspace}
\usepackage{booktabs}
\usepackage{todonotes}

\newcommand{\datasetname}{\textsc{FairytaleQA}\xspace}
\newcommand\tab[1][0.5cm]{\hspace*{#1}}

\definecolor{clabel}{HTML}{929292}

\definecolor{cinput}{HTML}{6EB4FD}
\definecolor{coutput}{HTML}{F89151}
\definecolor{cgroundtruth}{HTML}{99C893}

\title{\textit{Fantastic Questions and Where to Find Them}:\\ 
\datasetname \xspace-- An Authentic Dataset for Narrative Comprehension}

\author{
Ying Xu$^{\dagger}$\thanks{$^{\dagger}$Equal contributions \texttt{ying.xu@uci.edu, dakuo.wang@ibm.com, moyumyu@tencent.com, drritchi@uci.edu, yaob@rpi.edu. Work done while Mo was at IBM}. $^*$ Corresponding Author.} \\ \small{University of California Irvine}
\And Dakuo Wang$^{\dagger}$ \\ \small{IBM Research}
\And Mo Yu$^{\dagger}$ \\ \small{WeChat AI, Tencent}
\AND Daniel Ritchie$^{\dagger}$ \\ \small{University of California Irvine}
\And Bingsheng Yao$^{\dagger}$ \\ \small{Rensselaer Polytechnic Institute}
\And Tongshuang Wu \\ \small{University of Washington}
\AND Zheng Zhang \\ \small{University of Notre Dame}
\And Toby Jia-Jun Li \\ \small{University of Notre Dame}
\And Nora Bradford \\ \small{University of California Irvine}
\AND Branda Sun \\ \small{University of California Irvine}
\And Tran Bao Hoang \\ \small{University of California Irvine}
\And Yisi Sang \\ \small{Syracuse University}
\AND Yufang Hou \\ \small{IBM Research Ireland}
\And Xiaojuan Ma \\ \small{Hong Kong Univ. of Sci and Tech}
\And Diyi Yang \\ \small{Georgia Institute of Technology}
\AND Nanyun Peng \\ \small{University of California Los Angeles}
\And Zhou Yu \\ \small{Columbia University}
\And Mark Warschauer \\ \small{University of California Irvine}
}

\begin{document}
\maketitle
\begin{abstract}
Question answering (QA) is a fundamental means to facilitate assessment and training of narrative comprehension skills for both machines and young children, yet there is scarcity of high-quality QA datasets carefully designed to serve this purpose. In particular, existing datasets rarely distinguish fine-grained reading skills, such as the understanding of varying narrative elements. 
Drawing on the reading education research, we introduce \datasetname \footnote{Our dataset is available at \url{https://github.com/uci-soe/FairytaleQAData}.}, a dataset focusing on narrative comprehension of kindergarten to eighth-grade students. Generated by educational experts based on an evidence-based theoretical framework, \datasetname consists of 10,580 explicit and implicit questions derived from 278 children-friendly stories, covering seven types of narrative elements or relations. Our dataset is valuable in two folds: First, we ran existing QA models on our dataset and confirmed that this annotation helps assess models' fine-grained learning skills. Second, the dataset supports question generation (QG) task in the education domain. Through benchmarking with QG models, we show that the QG model trained on \datasetname is capable of asking high-quality and more diverse questions.

\end{abstract}

\newcommand{\labelstr}[1]{{ \color{clabel} \texttt{#1}}}
\section{Introduction}

\begin{table}[t!]
\centering
\fontsize{8.5}{9.5}\selectfont

\begin{tabular}{p{0.45\textwidth}}
    \toprule
    \textbf{Story Title:} \textit{\textbf{Magic Apples}} \\
    \textbf{Story Text:} \\
    \labelstr{[Sect 1]} Once upon a time there was a lad who was better off than all the others. He was never short of money, for he had a purse which was never empty. ... \\
    ... \\

    \labelstr{[Sect 6]} When the king's daughter had eaten of the apples, she had a pair of horns. And then there was such a wailing in the castle that it was pitiful to hear. ... \\
    \tab But one day a foreign doctor from afar came to court. He was not from their country, he said, and had made the journey purposely just to try his luck here. But he must see the king's daughter alone, said he, and permission was granted him. \\
    ... \\
    \labelstr{[Sect 8]} ... \\

    \midrule
    \begin{tabular}{@{\labelitemi\hspace{\dimexpr\labelsep+0.5\tabcolsep}}p{.43\textwidth}l@{}}\small{\textcolor{cinput}{ \textbf{Q1}}:Who will the foreign doctor turn out to be?}\end{tabular}                \\
    \labelstr{[explicit][prediction][sect 5, sect 6]} \\
    \begin{tabular}{@{\labelitemi\hspace{\dimexpr\labelsep+0.5\tabcolsep}}p{.43\textwidth}l@{}}\small{\textcolor{coutput}{ \textbf{A}}: The Lad. }\end{tabular}                                                \\ 
    
    \midrule
    
    \begin{tabular}{@{\labelitemi\hspace{\dimexpr\labelsep+0.5\tabcolsep}}p{.43\textwidth}l@{}}\small{\textcolor{cinput}{ \textbf{Q2}}:How did the princess feel when she had a pair of horns?}\end{tabular}                        \\
    \labelstr{[implicit][feeling][sect 6]} \\
    \begin{tabular}{@{\labelitemi\hspace{\dimexpr\labelsep+0.5\tabcolsep}}p{.43\textwidth}l@{}}\small{\textcolor{orange}{ \textbf{A}}: Upset. }\end{tabular}\\ \begin{tabular}{@{\labelitemi\hspace{\dimexpr\labelsep+0.5\tabcolsep}}p{.43\textwidth}l@{}}\small{\textcolor{orange}{ \textbf{A}}: Angry. }\end{tabular}\\
    \begin{tabular}{@{\labelitemi\hspace{\dimexpr\labelsep+0.5\tabcolsep}}p{.43\textwidth}l@{}}\small{\textcolor{orange}{ \textbf{A}}: Horrified. }\end{tabular}\\
   
    \bottomrule
    
    \end{tabular}
    \caption{\small{Story and Question-Answer examples in \datasetname. Each question has meta info (implicitness, question type, and section origin), and may have multiple answers and span across multiple sections.}}
    \label{tab:action_ex}
    \vspace{-1em}
\end{table}

Reading comprehension is a complex, multidimensional cognitive process~\cite{kim2017simple}. Question answering (QA) is fundamental for supporting humans' development of reading comprehension skills, as questions serve as both instruments for evaluation and tools to facilitate learning. To achieve this goal, comprehension questions should be valid and reliable, meaning that all items are designed to cohesively assess comprehension rather than some other skills (e.g., text matching, paraphrasing, or memorization) ~\cite{roberts2006reliability}. Moreover, from the educational perspective, given that reading comprehension is a multi-component skill, it is ideal for comprehension questions to be able to identify students' performance in specific sub-skills, thus allowing teachers to provide tailored guidance ~\cite{francis2005dimensions}. 

However, creating a large and suitable set of questions for supporting narrative comprehension is both time-consuming and cognitively demanding. Some researchers have proposed developing models to automatically generate questions or QA-pairs that satisfy the need for a continuous supply of new questions ~\cite{kurdi2020systematic, yao2022storybookqag}, which can potentially enable large-scale development of AI-supported interactive platforms for the learning and assessment of reading comprehension skills (e.g., ~\cite{zhang_storybuddy:_2022}). However, existing datasets are not particularly suitable for training question generation (QG) models for educational purposes ~\cite{das2021automatic}. This is primarily because the datasets are not typically structured around the specific dimensions of reading comprehension sub-skills, nor do they provide sufficient information on what sub-skills are tested. Consequently, QG models built on these datasets only yield one single ``comprehension'' score without a more detailed breakdown of performance on comprehension sub-skills. This issue is compounded by the fact that many benchmarks rely on crowd-sourced workers who may not have sufficient training or education domain knowledge needed to create valid questions in a consistent way. 

To bridge the gap, we constructed  \datasetname, an open-source dataset focusing on comprehension of narratives, targeting students from kindergarten to eighth grade. We focus on narrative comprehension for two reasons. First, narrative comprehension is a high-level comprehension skill strongly predictive of reading achievement ~\cite{lynch2008development} and plays a central role in daily life as people frequently encounter narratives in different forms ~\cite{goldie2003one}. Second, narrative stories have a clear structure of specific elements and relations among these elements, and there are existing validated narrative comprehension frameworks around this structure, which provides a basis for developing the annotation schema for our dataset.

We employed education experts who generated 10,580 question-answer pairs based on a collection of 278 fairytale stories for young readers, following evidence-based narrative comprehension frameworks ~\cite{paris2003assessing, alonzo2009they}. Thereby, \datasetname contains questions that focus on several narrative elements and relations, increasing the validity and reliability of the assessment. In addition, \datasetname also contains both explicit questions that involve answers found explicitly in the text and implicit questions that require high-level summarization (Table~\ref{tab:action_ex}), thus representing a relatively balanced assessment with questions of varying difficulty ~\cite{zucker2010preschool,raphael1986teaching}. Most importantly, our selection of annotators with education domain knowledge as well as the training and quality control process ensured that the aforementioned annotation protocol was consistently implemented. A subset of questions in our dataset has been validated with 120 pre-kindergarten and kindergarten students (IRB approved from the first author's institution), proving the questions' reliability and validity.  

We show the utility of \datasetname through two benchmarking experiments.
First, we used our data to train and evaluate state-of-the-art (SOTA) QA models and demonstrated that (1) \datasetname contains challenging phenomena for existing models, and (2) it can support finer-grained analysis on the different types of comprehension sub-skills, even for models trained on general QA datasets (NarrativeQA~\cite{kovcisky2018narrativeqa}).
We further calibrated model performances with human baseline, highlighting the most visible gap in models' reasoning capabilities on recognizing casual relationships and predicting event outcomes.
Second, we used \datasetname to power question generation and showed that the QG model trained on ours was more capable of asking diverse questions and generating questions with higher quality.

\section{Related Work}

\begin{table*}[t]
\small
\centering
\resizebox{\textwidth}{!}{%
\begin{tabular}{lccccccc}
\toprule
\textbf{Dataset}     & \textbf{Educ.} & \textbf{Narr.} & \textbf{Q. Type} & \textbf{A. Type} & \textbf{A. Source} & \textbf{Generation} & \textbf{Document Source}     \\
\midrule
\textbf{NarrativeQA} & No & Yes & Open-ended & Natural & Free-form & Crowd-sourced & {\begin{tabular}[c]{@{}c@{}}Movie Scripts, Literature\\ \scriptsize{(Full story or summary)}\end{tabular}} \\
\textbf{BookTest} & No & Yes & Cloze & Mult. Choice & Entity/Span & Automated & {\begin{tabular}[c]{@{}c@{}}Literature\\ \scriptsize{(Excerpt)}\end{tabular}} \\

\textbf{TellMeWhy} & No & Yes & Open-ended & Natural &  Free-form & Crowd-sourced & Short Fiction (ROCStories) \\

\textbf{RACE} & Yes & No & Open-ended & Mult. Choice & Free-form & Expert & {\begin{tabular}[c]{@{}c@{}}(Partially) Literature\\ \scriptsize{(Short story or excerpt)}\end{tabular}} \\
\textbf{CLOTH} & Yes & No & Cloze & Mult. Choice & Span & Expert & {\begin{tabular}[c]{@{}c@{}}(Partially) Literature\\ \scriptsize{(Short story or excerpt)}\end{tabular}} \\
\midrule
\textbf{\datasetname} & Yes & Yes & Open-ended & Natural & Free-form \& Span & Expert & {\begin{tabular}[c]{@{}c@{}}Literature\\ \scriptsize{(Full story)}\end{tabular}} \\
\bottomrule
\end{tabular}
}
\caption{Properties of existing datasets compared to \datasetname. }
\label{tab:existing_ds}
\end{table*}


\subsection{QA Datasets Focusing on Narratives}
Despite a large number of datasets on reading comprehension, few focus on comprehension of narrative text. Table~\ref{tab:existing_ds} reviews different narrative-related properties of existing popular QA datasets comparing with our proposed \datasetname dataset.
NarrativeQA ~\cite{kovcisky2018narrativeqa} is one of the representative datasets. It was generated by crowd workers who wrote QA pairs according to summaries of books or movie scripts, while the task takers are supposed to answer these questions based on their reading of original books or movie scripts. As such, this dataset is posited to evaluate a person's understanding of the underlying narrative, with a significant amount of event-related questions~\cite{mou2021narrative}. However, NarrativeQA simply instructed crowd-sourced workers to generate questions as if they were to ``test students'' without using a detailed annotation protocol. It is questionable whether these workers actually had experiences in testing students, and the lack of protocol may have imposed too little control over the coverage of reading sub-skills. 

BookTest~\cite{bajgar2016embracing} is an automatically constructed cloze-style QA dataset based on a collection of narrative texts retrieved from Project Gutenberg. The questions were generated by automatically removing a noun or entity in a sentence that has appeared in the preceding context. While cloze-style tests can be a valid instrument for assessing reading comprehension, their validity depends on the careful selection of words to be removed so that filling them in requires proper comprehension ~\cite{gellert2013cloze}. It is unlikely that automatically constructed cloze tests would meet such standards. 

Another dataset, TellMeWhy~\cite{lal2021tellmewhy}, aims to facilitate and assess understanding of \textit{causal relationships}. This dataset contains ``why'' questions that are relatively challenging, given that they require additional information not directly provided in the text. However, TellMeWhy only addresses one narrative component type (i.e., causal relationship), whereas \datasetname provides seven evaluation components. Moreover, TellMeWhy was built upon ROCStories~\cite{mostafazadeh2016corpus} and thus only examine comprehension on incomplete story exerpts, which may have limited the dataset's ability to assess macro-level summarization and inference making. 

\subsection{QA Datasets for Reading Education}
There are several benchmarks derived from sources for education purposes (e.g., exams or curricula). RACE~\cite{lai2017race} is a large-scale dataset consisting of comprehension questions from English exams for Chinese middle and high school students. RACE uses a mixture of narrative and informational paragraphs. These two genres require slightly different comprehension skills ~\cite{liebfreund2021cognitive}, and students perform differently based on what genres of text they read ~\cite{denton2015text}. Mixing these two together in one dataset without annotating the specific genre of each story/question obscures the ability to offer a precise assessment. Moreover, RACE is in multiple-choice format, and paragraphs are usually shorter. These two characteristics may make the RACE dataset less challenging, and recent models have demonstrated close-to-human performance\footnote{\url{http://www.qizhexie.com/data/RACE_leaderboard.html}}.

CLOTH~\cite{xie2017large} is a cloze-style dataset also collected from English exams with multiple choice fill-in-the-blank questions. CLOTH can be advantageous for educational QG as each question is labeled with the level of reasoning it involves. However, this dataset shares certain limitations inherent to multiple-choice formats ~\cite{klufa2015multiple}.

\subsection{Non-QA Datasets for Narrative Comprehension}
There are some datasets that are designed for assessing narrative comprehension skills but do not use QA as a form of evaluation. Several datasets, such as NovelChapters~\cite{ladhak2020exploring} and BookSum ~\cite{kryscinski2021booksum}, evaluate models' comprehension through summarization tasks. However, there have been debates of whether comprehension can be assessed solely through summarization ~\cite{head1989examination}, as summarization poses a high demand on writing that confounds the reading skills intended to be assessed.  
Two other recent datasets focus on singular specific elements in narratives. The LiSCU dataset~\cite{brahman2021let} targets readers' understanding of \textit{characters}, and ~\citet{sims2019literary} propose a dataset for detecting \textit{events} in narratives. Given their focus on single narrative elements, these two datasets may not provide a comprehensive evaluation of narrative comprehension.

\section{\datasetname }

\begin{table*}[!ht]
\centering
\resizebox{\textwidth}{!}{%

\begin{tabular}{c||c|c|c|c||c|c|c|c||c|c|c|c} 
\toprule
\multirow{3}{*}{\textbf{\begin{tabular}[c]{@{}c@{}} \datasetname \\ Dataset \end{tabular}}}   &   \multicolumn{4}{c||}{\textbf{Train} }   &   \multicolumn{4}{c||}{\textbf{Validation} }   &   \multicolumn{4}{c}{\textbf{Test} }  \\
\cmidrule{2-13}
      &   \multicolumn{4}{c||}{232 Books with 8548 QA-pairs}   &   \multicolumn{4}{c||}{23 Books with 1025 QA-pairs}  
      &   \multicolumn{4}{c}{23 Books with 1007 QA-pairs}   \\ 
\cmidrule{2-13}       
      &   \textbf{Mean}   &    \textbf{S.D.}   &    \textbf{Min}   &    \textbf{Max}   &    \textbf{Mean}   &    \textbf{S.D.}   &    \textbf{Min}   &    \textbf{Max}   &    \textbf{Mean}   &    \textbf{S.D.}   &    \textbf{Min}   &    \textbf{Max}   \\

\midrule\midrule
\# section per story   &   14.4   &   8.8   &   2   &   60   &   16.5   &   10.0   &   4   &   43   &   15.8   &   10.8   &   2   &   55   \\ 
\# tokens per story   &   2160.9   &   1375.9   &   228   &   7577   &   2441.8   &   1696.9   &   425   &   5865   &   2313.4   &   1369.6   &   332   &   6330   \\ 
\# tokens per section   &   149.6   &   64.8   &   12   &   447   &   147.8   &   56.7   &   33   &   298   &   145.8   &   58.6   &   24   &   290   \\ 
\# questions per story   &   36.8   &   28.9   &   5   &   161   &   44.5   &   29.5   &   13   &   100   &   43.7   &   28.8   &   12   &   107   \\ 
\# questions per section   &   2.8   &   2.440   &   0   &   18   &   2.9   &   2.3   &   0   &   16   &      3.0   &   2.4   &   0   &   15   \\ 
\# tokens per question   &   10.2   &   3.2   &   3   &   27   &   10.9   &   3.2   &   4   &   24   &   10.5   &   3.1   &   3   &   25   \\ 
\# tokens per answer   &   7.1   &   6.0   &   1   &   69   &   7.7   &   6.3   &   1   &   70   &   6.8   &   5.2   &   1   &   44   \\
\bottomrule
\end{tabular}
}

\medskip
\vspace{-10pt}
\caption {\small Core statistics of the \datasetname dataset, which has 278 books and 10580 QA-pairs.}
\vspace{-1em}
\label{tab:stats_fairytale}
\end{table*}

We developed the \datasetname dataset to address some of the limitations in existing benchmarks. Our dataset contains 10,580 QA pairs from 278 classic fairytale stories. In the remainder of this section, we report the dataset construction process and its key statistics.

\subsection{Source Texts}
The narrative texts utilized in the dataset are classic fairytales with clear narrative structures. We gathered the text from the Project Gutenberg website\footnote{\url{https://www.gutenberg.org/}}, using ``\emph{fairytale}'' as the search term. Due to a large number of fairytales found, we used the most popular stories based on the number of downloads since these stories are presumably of higher quality. 

To ensure the readability of the text, we made a small number of minor revisions to some obviously outdated vocabulary (e.g., changing ``ere'' to ``before'') and the unconventional use of punctuation (e.g., changing consecutive semi-colons to periods). For each story, we evaluated the reading difficulty level using the textstat\footnote{\url{https://pypi.org/project/textstat/}} Python package, primarily based on sentence length, word length, and commonness of words. We excluded stories that are at 10th grade level or above. 

These texts were broken down into small sections based on their semantic content by our annotators. The annotators were instructed to split the story into sections of 100-300 words that also contain meaningful content and are separated at natural story breaks. An initial annotator would split the story, and this would be reviewed by a cross-checking annotator. Most of the resulting sections were one natural paragraph of the original text. However, sometimes several paragraphs were combined (usually dialogue); and some exceptionally long paragraphs that contained more than one focal event were divided into multiple sections. On average, there are 15 sections per story, and each section has an average of 150 words (Table~\ref{tab:data_stats}).

\subsection{Schema for Question Annotation}
\label{ssec:schema}
\paragraph{Categorization via Narrative Elements or Relations}
\datasetname is intended to include QA pairs that capture the seven narrative elements/relations that are verified in prior educational research~\cite{paris2003assessing}. Definitions of question types are shown below. Example questions for each type are in Appendix \ref{appendix:example_qs}.

\definecolor{shadecolor}{rgb}{0.92,0.92,0.92}
\definecolor{Gray}{gray}{0.95}
\begin{itemize}[nosep, leftmargin=1em,labelwidth=*,align=left]
  \item \textbf{Character} questions ask test takers to identify the character of the story or describe characteristics of characters. 
  
  \item \textbf{Setting} questions ask about a place or time where/when story events take place and typically start with ``\emph{Where}'' or ``\emph{When}''.
  
  \item \textbf{Action} questions ask about characters' behaviors or information about that behavior.
  
  \item \textbf{Feeling} questions ask about the character's emotional status or reaction to certain events and are typically worded as ``\emph{How did/does/do …feel}''. 
  
  \item \textbf{Causal relationship} questions focus on two events that are causally related where the prior events causally lead to the latter event in the question. This type of questions usually begins with ``\emph{Why}'' or ``\emph{What made/makes}''. 
  
  \item \textbf{Outcome resolution} questions ask for identifying outcome events that are causally led to by the prior event in the question. This type of questions are usually worded as ``\emph{What happened/happens/has happened...after...}''. 
  \item \textbf{Prediction} questions ask for the unknown outcome of a focal event, which is predictable based on the existing information in the text. 
\end{itemize}

These labels are to ensure the presence of the variety of questions' sub-skills so that the models trained on this dataset can also generate the variety. The labels are not intended to aid the training of a model to classify questions. Some -- but not all -- of the labels may be determined by surface features. For example, \textit{feeling} questions typically contain the words ``feel'' or ``feels'', while  \textit{action} questions are more broad in their format.

\paragraph{Categorization via Source of Answers}
Orthogonal to the aforementioned question categories, questions in \datasetname are also categorized based on whether or not the answer source can be directly found in the text, namely explicit versus implicit questions. In general, explicit questions revolve around a specific story fact, and implicit questions require summarizing and making an inference based on information that is only implicit in the text. Using a combination of explicit and implicit questions yields an assessment with more balanced difficulty ~\cite{raphael1986teaching, zucker2010preschool}. In our data, explicit and implicit questions are defined as below (Examples in Appendix C): 

\begin{itemize}[nosep, leftmargin=1em,labelwidth=*,align=left]
  \item \textbf{Explicit} questions ask for answers that can be directly found in the stories. In other words, the source of answer are spans of text. 
   
  \item \textbf{Implicit} questions ask for answers that cannot be directly found in the text. Answering the questions require either reformulating language or making inference. In other words, the answer source is ``free-form'', meaning that the answers can be any free-text, and there is no limit to where the answer comes from. 

\end{itemize}

\subsection{Annotation Process}
Five annotators were involved in the annotation of QA pairs. All of these annotators have a B.A. degree in education, psychology, or cognitive science and have substantial experience in teaching and reading assessment. These annotators were supervised by three experts in literacy education. 

\paragraph{Annotation Guidelines}
The annotators were instructed to imagine that they were creating questions to test elementary or middle school students in the process of reading a complete story. We required the annotators to generate only natural, open-ended questions 
, avoiding ``yes-'' or ``no-'' questions. We also instructed them to provide a diverse set of questions about 7 different narrative elements, and with both implicit and explicit questions. Each question in the dataset has a label on the narrative element/relation to be assessed and whether it is implicit or explicit.  

We asked the annotators to also generate answers for each of their questions. We asked them to provide the shortest possible answers but did not restrict them to complete sentences or short phrases. For explicit questions, annotators extracted the shortest phrase from the text as the answer (i.e., span). For implicit questions, annotators provided at least two possible answers for each question (i.e., free-form). We also asked the annotators to label which section(s) the question and answer was from. We did not specify the number of questions per story to account for story length variability and to allow annotators to create meaningful questions rather than be forced to add unnecessary questions. However, we did ensure that the annotators broadly averaged 2-3 questions per \textit{section} in order to guarantee dataset size.

\paragraph{Annotator Training and Cross-Checking}
All annotators received a two-week training in which each of them was familiarized with the coding template (described in the section below) and conducted practice coding on the same five stories. The practice QA pairs were then reviewed by the other annotators and the three experts, and discrepancies among annotators were discussed. At the end of the training session, the five annotators had a little disagreement with the questions generated by other coders. During the annotation process, the team met once every week to review and discuss each member’s work. All QA pairs were cross-checked by two annotators, and 10\% of the QA pairs were additionally checked by the expert supervisor. This process was to ensure that the questions focused on key information to the narrative and the answers to the questions were correct. 

\begin{table}[]
\centering
\small
\begin{tabular}{lrrrr}
\toprule
& \textbf{Mean} & \textbf{Min} & \textbf{Max}  & \textbf{SD}    \\ \midrule\midrule
\multicolumn{5}{l}{\textbf{Story Characteristics}}          \\
Sections / story    & 14.7   & 2   & 60   & 9.2  \\
Tokens / story      & 2196.7 & 228 & 7577 & 1401.3  \\
Tokens / section    & 149.1   & 12  & 447  & 63.6 \\ \midrule
\multicolumn{5}{l}{\textbf{Question Characteristics}}          \\
Tokens / question   & 10.3   & 3   & 27   & 3.3  \\
Tokens / answer     & 7.2    & 1   & 69   & 6.1  \\
Questions / story   & 38.1   & 5   & 161  & 29    \\
Questions / section & 2.9    & 0   & 18   & 2.4  \\
\bottomrule
\end{tabular}
\caption{\small{Various descriptive statistics for the length of stories and number of questions in the dataset.}}
\label{tab:data_stats}
\end{table}

\begin{table}[t]
\centering
\small
\begin{tabular}{lrr}
\toprule
\textbf{Category}   & \textbf{Count} & \textbf{Percentage (\%)} \\ \midrule\midrule
\multicolumn{3}{l}{\textbf{Attributes}}                    \\
character           & 1172           & 11.08             \\
causal relationship & 2940           & 27.79             \\
action              & 3342           & 31.59             \\
setting             & 630            & 5.95              \\
feeling             & 1024           & 9.68              \\
prediction          & 486            & 4.59              \\
outcome resolution  & 986            & 9.32              \\
\midrule
\multicolumn{3}{l}{\textbf{Explicit vs Implicit}}          \\
explicit            & 7880           & 74.48             \\
implicit            & 2700           & 25.52             \\
\bottomrule
\end{tabular}
\caption{\small{Breakdown of questions per category based on the schema in Section~\ref{ssec:schema}.}} 
\label{tab:category_vals}
\end{table}

\paragraph{Agreement among Annotators} The questions generated by the five coders showed a consistent pattern. All coders' questions have similar lengths (average length ranging from 8 to 10 words among the coders) and have similar readability levels (average readability between fourth to fifth grade among the coders). The distributions in narrative elements focused as well as implicit/explicit questions were also consistent. A detailed description of the distributions by coders is displayed in Appendix \ref{app:att_plot}. We chose not to use traditional inter-annotator agreement (IAA) metrics like Kappa coefficients because we explicitly asked the coders to generate questions and answers with variable language to aid QA and QG models based on this dataset. This language variability leads to inaccurate IAA metrics by traditional means~\cite{amidei2018kappa}, leading to our decision.

\paragraph{Second Answer Annotation}
For the 46 stories used as the evaluation set, we annotate a second reference answer by asking an annotator to independently read the story and answer the questions generated by others. All questions were judged as answerable and thus answered by the second annotator.
The second answers are used for both human QA performance estimation and for providing multiple references in automatic QA evaluation.

\subsection{Statistics of \datasetname }

We random split the \datasetname dataset into train/val/test splits with a QA ratio of roughly 8:1:1. Table~\ref{tab:stats_fairytale} shows the detailed statistics of the \datasetname Dataset in train/val/test splits.

Overall, the resulting \datasetname dataset contained 10,580 questions from 278 fairytale stories. The description of story and question characteristics is presented in Table~\ref{tab:data_stats}. In \datasetname, \textit{action} and \textit{causal relationship} questions are the two most common types, constituting 31.6\% and 27.8\%, respectively, of all questions.\textit{ Outcome resolution, character,} and \textit{feeling} types each constitute about 10\% of all questions. \textit{Setting} and \textit{prediction} questions are about 5\% each. Our dataset contains about 75\% explicit questions and 25\% implicit questions (Table \ref{tab:category_vals} for details). 

\paragraph{Validation of \datasetname for Comprehension Assessment}
We validated the questions in \datasetname using established procedures in educational assessment development ~\cite{ozdemir2019development} and have proven that our questions have high reliability and validity. Specifically, we sampled a small subset of the questions in our dataset (11 questions generated for one story) and tested them among 120 students in prekindergartens and kindergartens. This study was pre-approved by the IRB in first author's institution. The Cronbach's coefficient alpha was 0.83 for the items in this story comprehension assessment; suggesting was high internal reliability. We also linked children's performance answering our questions to another validated language assessment ~\cite{martin2011expressive}, and the correlation was strong 0.76 (p<.001), suggesting an excellent external validity.

\section{Baseline Benchmark: Question Answering}

In the following sections, we present a couple of baseline benchmarks on both the Question Answering (QA) task and the Question Generation (QG) task with \datasetname. 
We leveraged both pre-trained neural models and models fine-tuned on different QA datasets, including NarrativeQA and our dataset, \datasetname. The baseline results show that our \datasetname demonstrates challenging problems to existing approaches, and those models fine-tuned on \datasetname can benefit from the annotations a lot to achieve significant performance improvement. We also report human performance by scoring one reference answer to the other.   

\subsection{Question Answering Task and Model}

Question Answering (QA) is a straightforward task that our \datasetname dataset can contribute to. We leveraged the commonly-used Rouge-L F1 score for the evaluation of QA performances. For each QA instance, we compared the generated answer with each of the two ground-truth answers and took the higher Rouge-L F1 score. 

\begin{table}[]
\centering
\small
\resizebox{.48\textwidth}{!}{%
\begin{tabular}{lc}
\toprule
\textbf{Model}                                 & 
\multicolumn{1}{c}{\textbf{Validation / Test}} \\
                                               &
\multicolumn{1}{c}{\textbf{ROUGE-L F1}}      \\ 

\midrule
\textbf{Pre-trained Models}                    &                                              \\
BERT                                           &  0.104 / 0.097                                 \\
DistilBERT                                     &  0.097 / 0.082                                 \\
BART                                           &  0.108 / 0.088                                 \\
\midrule
\textbf{Fine-tuned Models}                     &                                              \\
BART fine-tuned on NarrativeQA                 & 0.475 / 0.492                                  \\
BART fine-tuned on \datasetname                    & 0.533 / 0.536                                  \\
\midrule
Human$^\ddagger$                                          & \textbf{0.651 / 0.644}                         \\ 
\bottomrule
\end{tabular}%
}
\caption{\small{Question Answering benchmarks on \datasetname validation and test splits. $\ddagger$Human results are obtained via cross-estimation between the two annotated answers, thus are underestimated. Still, they outperform models. We leave a full large-scale human study to future work.}}
\label{tab:qa_performance}
\end{table}

\subsection{Main Results}
Here in Table \ref{tab:qa_performance}, we show the QA performance of a few pretrained SOTA neural-model architectures: BERT~\cite{devlin2018bert}, BART~\cite{lewis2019bart}, and DistilBERT\cite{sanh2019distilbert}. The quality of answers generated by these pre-trained models is on par with each other. Since BART outperformed other model architectures in the QA task of NarrativeQA \cite{mou2021narrative}, we decided to use BART as the backbone for our fine-tuned models. 

We report the performance of fine-tuned BART models with the following settings: BART fine-tuned on NarrativeQA, which is the SOTA model reported in \cite{mou2021narrative}, and another BART model fine-tuned on \datasetname. We note that for the QA task, the model that was fine-tuned on \datasetname dataset performs much better than the model fine-tuned on NarrativeQA by at least 5\%. Even the human performance is underestimated here because it is obtained via cross-estimation between two annotated answers, this result still leaves around 12\% on both splits between human performance and the model fine-tuned with \datasetname, which demonstrates that the QA task is still a challenging problem for existing works on our \datasetname dataset. We leave a full large-scale human study for evaluating the accurate human performance to future work. 

\subsection{Analysis}

\paragraph{Performance Decomposition}

\begin{figure}[t]
    \includegraphics[width=0.47\textwidth]{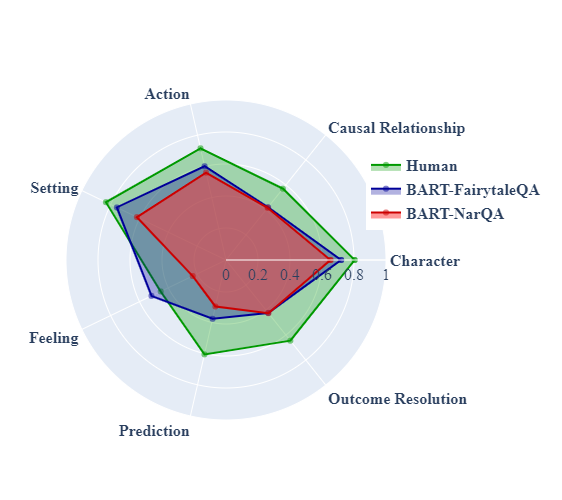}

\caption{\small{Decomposed QA results (Rouge-L) on 7 narrative elements on the validation split.}}
\label{fig:decompose_qa_result_val}
\end{figure}



%




Given that \datasetname has question type annotations on all the question-answer pairs, it supports the decomposition of performance on different types, thus resulting in a comprehensive picture of which reading skills the models lack the most.


Figure~\ref{fig:decompose_qa_result_val} presents the QA performance decomposition as a radar visualization. (The full results on both validation and test sets can be found in Table~\ref{tab:decompose_qa_result_appendix} in Appendix~\ref{app:decompose}). Compared to the model trained on NarrativeQA, our \datasetname led to the biggest improvement on dimensions of \texttt{Setting} and \texttt{Feeling} with more than 10\% increase.
The \texttt{Character} and \texttt{Prediction} dimensions were also improved by a large margin (7-8\%).
The large improvements in these dimensions suggested that despite the NarrativeQA dataset's overall focus on narrative comprehension, it might not include questions that sufficiently cover some of the fundamental elements, probably due to the lack of detailed annotating protocol and typical crowd workers' limited knowledge in reading assessment.

By comparison, on dimensions of \texttt{Action}, \texttt{Causal Relationship} and \texttt{Outcome Resolution}, our model fine-tuned on \datasetname resulted in smaller improvement compared to the model fine-tuned on NarrativeQA. This is likely due to the fact that most of the NarrativeQA questions are about event arguments and causal or temporal relations between events, as suggested by a human study \cite{mou2021narrative}.

Our performance decomposition also revealed substantial gaps between existing SOTA models and humans. Specifically, humans were 15-20\% better on \texttt{Causal Relationship}, \texttt{Outcome Resolution} and \texttt{Prediction}. The model-human performance gaps on \texttt{Causal Relationship} and \texttt{Outcome Resolution} likely reflected the deficiency of current NLP models in understanding story plots, and the gap on \texttt{Prediction} might be due to the fact that this dimension asked the models to envision what would come next in the text, which required connecting commonsense knowledge with the content of the text. The model-human performance gaps on \texttt{Character} and \texttt{Setting} were also considerable, suggesting that the models' ability to understand these basic reading elements still has much room for improvement. 

Finally, it was interesting that the model trained on our dataset outperformed humans on the \texttt{Feeling} dimension. This was likely because the answers to these \texttt{Feeling} questions were most explicitly described in the story. Therefore, it did not actually require reasoning of the character's mental states, but rather understanding which parts of the texts express the feelings. Another QA performance decomposition result based on explicit/implicit question types is provided in Appendix~\ref{app:decompose_ex_im}.

\begin{figure}[t]
    \centering
    \includegraphics[width=0.45\textwidth]{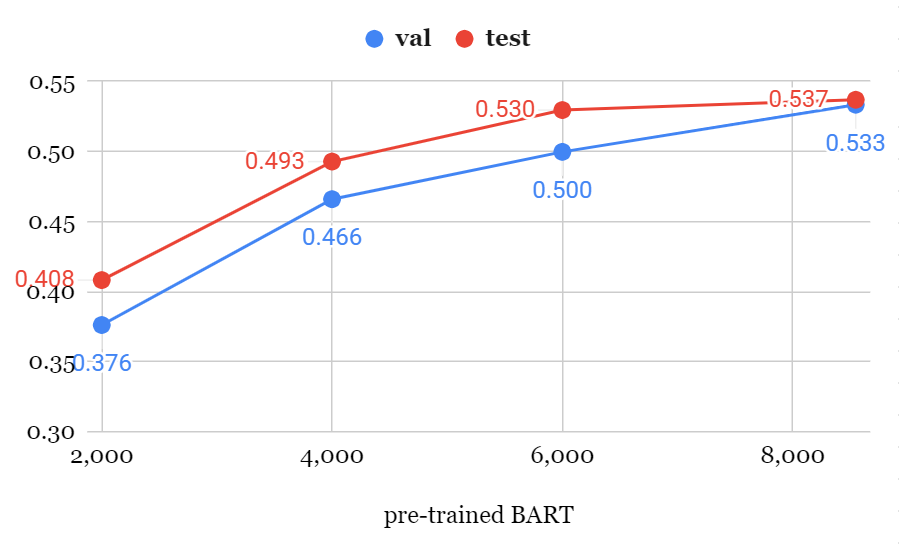}
    \vspace{-5pt}
    \caption{\small{Learning curve of the QA model on \datasetname with varying size of training data.}}
    \label{fig:learning_curve}
\end{figure}
\vspace{-5pt}

\paragraph{Learning Curve}
Finally, we present the learning curve of the BART QA model on our \datasetname. Figure~\ref{fig:learning_curve} plots the model performance on the validation set with different sizes of training data.
The curve became flatter after training with 6,000 QA pairs in our dataset. This suggested that our dataset has a reasonably good size for fine-tuning a SOTA pre-trained model, and the performance gap between models and humans requires a more sophisticated reading model design rather than solely augmenting the training examples.

\section{Baseline Benchmark: Question Generation}
\subsection{Question Generation Task and Model}

In terms of the QG performance on \datasetname, the task was to generate questions that correspond to the given answers and the context. 
This task has important empirical applications that in the future, models may help teachers to create questions in the educational settings.

Similar to the QA task, we fine-tuned a BART model to generate a question conditioned on each human-labeled answer and corresponding story section. The generated question is then evaluated with the corresponding ground-truth question. We used ROUGE-L F1 score as the evaluation metric. For this QG task, we compare the models fine-tuned on NarrativeQA, on \datasetname, and on both datasets. 



\begin{table}[t!]
\centering
\small
\resizebox{.48\textwidth}{!}{%
\begin{tabular}{lc}
\toprule
\textbf{Model}                                 & 
\multicolumn{1}{c}{\textbf{Validation / Test}} \\
                                               &
\multicolumn{1}{c}{\textbf{ROUGE-L F1}}      \\ 
\midrule

BART fine-tuned on NarrativeQA                 & 0.424 / 0.442                           \\
BART fine-tuned on \datasetname                    & \textbf{0.527 / 0.527}                  \\
BART fine-tuned on NarrativeQA and \datasetname    &  0.508 / 0.519                         \\ 
\bottomrule
\end{tabular}%
}
\caption{\small{Question Generation benchmarks on \datasetname-validation and test splits.}}
\label{tab:qg_performance}
\end{table}

\begin{table}[t]
\centering
\small
\resizebox{.48\textwidth}{!}{%
\begin{tabular}{l r r r}
\toprule
& \textbf{Groundtruth} & \begin{tabular}[c]{@{}c@{}} \textbf{BART-}\\ \textbf{NarQA} \end{tabular} &  \begin{tabular}[c]{@{}c@{}} \textbf{BART-}\\ \textbf{\datasetname} \end{tabular}       \\ 
\midrule
Who     & 84    & 62    & 97    \\
What    & 426   & 716   & 447   \\
Why     & 287   & 144   & 304   \\
How     & 178   & 59    & 129   \\
Where   & 44    & 35    & 47    \\
Other   & 6     & 9     & 1     \\

\bottomrule
\end{tabular}%
}
\caption{\small{Distribution of question word in QG task for validation split by benchmark models.}}
\label{tab:qg_stats}
\end{table}

\begin{table}[t]
\centering
\small
\resizebox{.48\textwidth}{!}{%
\begin{tabular}{p{0.45\textwidth}}
\toprule

\textbf{\textcolor{cinput}{Input story section:}}  the wild people who dwell in the south-west are masters of many black arts. they often lure men of the middle kingdom to their country by promising them their daughters in marriage, but their promises are not to be trusted. once there was the son of a poor family, who agreed to labor for three years for one of the wild men in order to become his son-in-law. \\
\midrule
\midrule
\textbf{\textcolor{cinput}{Input Answer 1: }} The son of a poor family.   \\
\midrule
\textbf{\textcolor{cgroundtruth}{Ground-truth Question}}                         \\
Who agreed to labor for three years for one of the wild men in order to become his son-in-law?  \\
\midrule
\textbf{\textcolor{coutput}{Outputs}}                            \\
\tab \textit{\textbf{BART-NarQA:}}  What was the son of a poor family?         \\
\tab \textit{\textbf{BART-\datasetname:}}  Who agreed to labor for one of the wild men in order become his son-in law?                                           \\
\midrule
\midrule
\textbf{\textcolor{cinput}{Input Answer 2: }} The wild people.   \\
\midrule
\textbf{\textcolor{cgroundtruth}{Ground-truth Question}}                         \\
Who dwelled in the south-west and were masters of many black arts?  \\
\midrule
\textbf{\textcolor{coutput}{Outputs}}                         \\
\tab \textit{\textbf{BART-NarQA:}}  What dwells in the south-west?         \\
\tab \textit{\textbf{BART-\datasetname:}}  Who dwell in the south-west are masters of many black arts?                                           \\

\bottomrule
\end{tabular}%
}

\caption{\small{Qualitative analysis of QG models fine-tuned on NarQA or FairytaleQA dataset.}}
\label{tab:qg_example}
\end{table}


\subsection{Results and Analysis}

Table \ref{tab:qg_performance} displays the QG results. 
The model fine-tuned on \datasetname demonstrated a clear advantage on Rouge-L over the model fine-tuned on NarrativeQA. It is worth noting that the model fine-tuned on both NarrativeQA and \datasetname performs worse than the model fine-tuned on \datasetname only; we would assume that NarrativeQA dataset introduces noises in question type distribution or semantics during the training process.

Further analysis (Table~\ref{tab:qg_stats}) examined the distribution of generated question types according to the beginning word of a question (wh- words). The questions generated by \datasetname more closely resembled the pattern of the ground-truth questions, suggesting that our dataset was able to improve the model's ability to mimic the education experts' strategy of asking questions that assess the seven elements of reading comprehension.

This result is further supported by qualitative analysis (as seen in examples in Table~\ref{tab:qg_example}). Compared to the QG model trained with \datasetname, the baseline model trained with NarrativeQA dataset tended to generate vague questions that did not build upon specific contextual evidence within the narratives. These kinds of vague questions may not be suitable in educational settings, as improving students' skills to find text evidence to support their comprehension is a crucial aspect of reading education. The disparity between the two models might be attributed to how the QA-pairs were constructed in these two datasets: while NarrativeQA was constructed by crowd workers who only read the abstract of the stories, \datasetname required annotators to read the complete story before developing QA-pairs. As such, it is not surprising that models trained on \datasetname dataset could generate questions that are more closely related to the contextual evidence within the original text. In addition, we also observed that the model trained on NarrativeQA tended to generate questions with seemingly more correct grammar but were factually inaccurate (Table~\ref{tab:qg_example_appeldix} Appendix~\ref{app:qg_example}).






\section{Conclusion and Future work}
In summary, we constructed a large-scale dataset, \datasetname, for children's narrative comprehension. Generated by educational domain experts, this dataset has been used to support preliminary work on QG tasks~\cite{yao2022storybookqag,zhao2022storybookqag} and has enabled downstream AI-for-Education applications~\cite{zhang_storybuddy:_2022,xu2021same}.

Our work has several limitations that warrant further exploration.
As discussed before, the human performance results for the QA task were underestimated because they were obtained via cross-estimation between the two annotated answers. One possibility for future work is to conduct a large-scale human annotation to collect more answers per question and then leverage the massively annotated answers to better establish a human performance evaluation. 

Another future direction is to leverage our dataset to identify social stereotypes represented in story narratives. It will be valuable to analyze how social stereotypes are represented in the children's literature collected in our dataset, thus enabling the development of automatic systems that detect and mitigate social biases. This type of bias analysis has been an underexplored research topic for the ML community, yet it will have profound societal impacts. 

These are two examples of the many new research and application opportunities potentially enabled by our \datasetname dataset, and we welcome researchers from both NLP and education communities to leverage \datasetname to advance NLP technologies and to promote education.


\section*{Acknowledgements}
We thank Schmidt Futures for providing funding for the development of the \texttt{FairytaleQA} dataset. This work is also supported by the National Science Foundation (Grant No. 1906321 and 2115382). 

\newpage

\bibliographystyle{acl_natbib}
\bibliography{anthology}

\clearpage

\appendix

\section{Decomposed QA results on 7 narrative elements for val/test splits}
\label{app:decompose}

\begin{table}[h]
\centering
\resizebox{.48\textwidth}{!}{%
\small

\begin{tabular}{lccc}
\toprule
& \textbf{\begin{tabular}[c]{@{}c@{}} BART-\\ NarQA \end{tabular}} 
& \textbf{\begin{tabular}[c]{@{}c@{}} BART-\\ \datasetname \end{tabular}}  
& \textbf{Human}    \\ 


\midrule
\multicolumn{4}{l}{\textbf{Validation}}          \\
Character           &   0.65    &	0.720   &	0.804\\
Causal Relationship &   0.417   &	0.422   &	0.570\\
Action              &   0.560   &	0.601   &	0.716\\
Setting             &   0.618   &	0.757   &	0.833\\
Feeling             &   0.231   &	0.517   &	0.453\\
Prediction          &   0.298   &	0.377   &	0.605\\
Outcome Resolution  &   0.425   &	0.423   &	0.645\\
\midrule
\multicolumn{4}{l}{\textbf{Test}}          \\
Character           &   0.691   &	0.757   &	0.864\\
Causal Relationship &   0.447   &	0.432   &	0.589\\
Action              &   0.559   &	0.608   &	0.710\\
Setting             &   0.683   &	0.696   &	0.755\\
Feeling             &   0.301   &	0.508   &	0.533\\
Prediction          &   0.275   &	0.300   &	0.366\\
Outcome Resolution  &   0.409   &	0.486   &	0.574\\

\bottomrule
\end{tabular}
}
\caption{Decomposed QA results on 7 narrative elements. }
\label{tab:decompose_qa_result_appendix}
\end{table}

\noindent Table~\ref{tab:decompose_qa_result_appendix} shows the full decomposed QA results on 7 narrative elements for both validation and test splits, in terms of BART fine-tuned on NarrativeQA, BART fine-tuned on \datasetname, and human performance for the experts created ground-truth QA-pairs.

\section{Decomposed QA results on explicit/implicit question types for val/test splits}
\label{app:decompose_ex_im}

\begin{table}[h]
\centering
\small

\resizebox{.48\textwidth}{!}{%
\begin{tabular}{lcc}
\toprule
\textbf{Model}                                 & 
\multicolumn{2}{c}{\textbf{Validation / Test}} \\
                                               &
\multicolumn{2}{c}{\textbf{ROUGE-L F1}}      \\ 
\midrule
& \textbf{Implicit} & \textbf{Explicit} \\
\midrule

BART-NarQA                 &    0.280/0.278    &    0.548/0.563                  \\
BART-\datasetname          &    0.304/0.286    &    0.619/0.620       \\
Human                      &    0.363/0.330    &    0.760/0.750     \\

\bottomrule
\end{tabular}%
}

\caption{\small{Decomposed QA results on implicit/explicit types.}}
\label{tab:decompose_qa_result_im_ex}
\end{table}

\noindent We provided another QA performance decomposition based on explicit/implicit question types in Table~\ref{tab:decompose_qa_result_im_ex}. We noticed that the implicit questions are much more difficult for both humans and models to answer, which only achieves roughly half the performance compared with explicit questions. This result is consistent with our expectation, where answers to explicit questions can be directly found in the content while implicit questions require high-level summarization and inference.

\section{QG examples by benchmark models on event-based answers}
\label{app:qg_example}

\begin{table}[t!]
\centering
\small
\resizebox{.48\textwidth}{!}{%
\begin{tabular}{p{0.45\textwidth}}
\toprule

\textit{\textbf{\textcolor{blue}{Input story section:}}}  you see from this that the sparrow was a truthful bird, and the old woman ought to have been willing to forgive her at once when she asked her pardon so nicely. but not so.the old woman had never loved the sparrow, and had often quarreled with her husband for keeping what she called a dirty bird about the house, saying that it only made extra work for her. now she was only too delighted to have some cause of complaint against the pet. she scolded and even cursed the poor little bird for her bad behavior, and not content with using these harsh, unfeeling words, in a fit of rage she seized the sparrow-who all this time had spread out her wings and bowed her head before the old woman, to show how sorry she was-and fetched the scissors and cut off the poor little bird's tongue. \\
\midrule
\midrule
\textit{\textbf{\textcolor{blue}{Input Answer: }}} Cut off the poor little bird's tongue.   \\
\midrule
\textit{\textbf{\textcolor{green}{Ground-truth Question}}}                         \\
What did the woman do to punish the bird?  \\
\midrule
\textit{\textbf{\textcolor{orange}{Outputs}}}                            \\
\tab \textit{\textbf{BART-NarQA:}}  What did the old woman do in her rage? \\
\tab \textit{\textbf{BART-\datasetname:}}  What did the old woman do after she seized her sparrow?                                 \\
\midrule
\midrule
\textit{\textbf{\textcolor{blue}{Input story section:}}}  ``do not be sparing of the silver pieces in your pocket!'' she cried after him as he went off.he went to the village, attended to everything, and came back. the woman tore the cloth apart, made a coat of it and put it on. no sooner had they walked a few miles before they could see a red cloud rising up in the south, like a flying bird.``that is my mother,'' said the woman.in a moment the cloud was overhead. then the woman took the black tea-cups and threw them at it. seven she threw and seven fell to earth again. and then they could hear the mother in the cloud weeping and scolding, and thereupon the cloud disappeared.they went on for about four hours. then they heard a sound like the noise of silk being torn, and could see a cloud as black as ink, which was rushing up against the wind.``alas, that is my father!'' said the woman. ``this is a matter of life and death, for he will not let us be! because of my love for you i will now have to disobey the holiest of laws!'' \\
\midrule
\textit{\textbf{\textcolor{blue}{Input Answer: }}} Took the black tea-cups and threw them at it.   \\
\midrule
\textit{\textbf{\textcolor{green}{Ground-truth Question}}}                         \\
What did the wife do when she saw her mother?  \\
\midrule
\textit{\textbf{\textcolor{orange}{Outputs}}}                            \\
\tab \textit{\textbf{BART-NarQA:}}  What did the woman do to try and kill her father?         \\
\tab \textit{\textbf{BART-\datasetname:}}  What did the woman do after she saw her mother?     \\

\bottomrule
\end{tabular}%
}

\caption{Question Generation examples with event-related input answers by benchmark models. }
\label{tab:qg_example_appeldix}
\end{table}

\begin{table*}[!htb]
\small
\centering
\resizebox{\textwidth}{!}{%
\begin{tabular}{ll}
\toprule
\textbf{Category} & \textbf{Example QA Pair}\\
\midrule
\vspace{1.5mm}
\textbf{Character} & \begin{tabular}{p{13cm}}
    \rowcolor{Gray} \textcolor{red}{Q}: \textit{How did the man's daughter look?} \\
    \rowcolor{Gray} \textcolor{green}{A}: \textit{beautiful}\\
    \\
    \rowcolor{Gray}  \textcolor{red}{Q}: \textit{Who were the brother and sister living with after their mom died?} \\
    \rowcolor{Gray}  \textcolor{green}{A}: \textit{their stepmother}  
  \end{tabular} \\ \vspace{1.5mm}
\textbf{Setting} & \begin{tabular}{p{13cm}}
    \rowcolor{Gray} \textcolor{red}{Q}: \textit{Where did the man and his wife and two girls live?}\\
    \rowcolor{Gray} \textcolor{green}{A}: \textit{near the forest}  
  \end{tabular} \\ \vspace{1.5mm}
\textbf{Action} & \begin{tabular}{p{13cm}}
  \rowcolor{Gray} \textcolor{red}{Q}: \textit{What did the cook do after she opened the hamper?} \\
  \rowcolor{Gray} \textcolor{green}{A}: \textit{unpacked the vegetables}\\
  \\
  \rowcolor{Gray}  \textcolor{red}{Q}: \textit{How did Johnny Town-Mouse and his friends treat Timmy Willie when they met him?} \\
  \rowcolor{Gray} \textcolor{green}{A}: \textit{Johnny Town-Mouse and his friends treat Timmy Willie poorly.}
  \end{tabular} \\ \vspace{1.5mm}
\textbf{Causal relationship} & \begin{tabular}{p{13cm}}
  \rowcolor{Gray} \textcolor{red}{Q}: \textit{Why did the two mice come tumbling in, squeaking, and laughing?} \\
  \rowcolor{Gray} \textcolor{green}{A}: \textit{They were being chased by the cat.}
  \end{tabular} \\ \vspace{1.5mm}
\textbf{Outcome resolution} & \begin{tabular}{p{13cm}}
  \rowcolor{Gray} \textcolor{red}{Q}: \textit{What happened to Timmy after he got in the hamper?} \\
  \rowcolor{Gray} \textcolor{green}{A}: \textit{The hamper takes him to the garden.}
  \end{tabular} \\ \vspace{1.5mm}
\textbf{Feeling} & \begin{tabular}{p{13cm}}
  \rowcolor{Gray} \textcolor{red}{Q}: \textit{How did the princess feel in her new home?} \\
  \rowcolor{Gray} \textcolor{green}{A}: \textit{happy}
  \end{tabular} \\ \vspace{1.5mm}
\textbf{Prediction} & \begin{tabular}{p{13cm}}
  \rowcolor{Gray} \textcolor{red}{Q}: \textit{How will the other animals treat the duckling?} \\
  \rowcolor{Gray} \textcolor{green}{A}: \textit{The other animals will look down on the duckling.}
  \end{tabular} \\
\midrule
  \vspace{1.5mm}
\textbf{Explicit} & 
  \begin{tabular}{p{13cm}}
  \rowcolor{Gray} \textcolor{red}{Q}: \textit{How did the girl feel when she saw the old woman's teeth?} \\
  \rowcolor{Gray} \textcolor{green}{A}: \textit{terrified}\\
  \rowcolor{Gray} \textcolor{blue}{Context}: \textit{...but she had such great teeth that the girl was \textcolor{blue}{terrified}...}\\
  \\
  \rowcolor{Gray} \textcolor{red}{Q}: \textit{What happened when the door of the stove was opened?} \\
  \rowcolor{Gray} \textcolor{green}{A}: \textit{The flames darted out of its mouth.}\\
  \rowcolor{Gray} \textcolor{blue}{Context}: \textit{...when the door of the stove was opened, \textcolor{blue}{the flames darted out of its mouth}. This is customary with all stoves...}
  \end{tabular} \\ \vspace{1.5mm}
\textbf{Implicit} & \begin{tabular}{p{13cm}}
  \rowcolor{Gray} \textcolor{red}{Q}: \textit{What happened when the prince broke open one of the crow's eggs?}\\
  \rowcolor{Gray} \textcolor{green}{A1}: \textit{The prince found a beautiful palace inside.}\\
  \rowcolor{Gray} \textcolor{green}{A2}: \textit{There was a beautiful palace inside. }\\
  \rowcolor{Gray} \textcolor{green}{A3}: \textit{A little palace was inside and it grew until it covered as much ground as seven large barns.}\\
  \rowcolor{Gray} \textcolor{blue}{Context}: \textit{The Swan Maiden lit in a great wide field, and there she told the prince to break open one of the crow's eggs. The prince did as she bade him, and what should he find but the most beautiful little palace, all of pure gold and silver. He set the palace on the ground, and it grew and grew and grew until it covered as much ground as seven large barns.}
  \end{tabular} \\
\bottomrule
\end{tabular}
}
\caption{Example QA-pairs of \datasetname. We show one QA-pair for each narrative element as well as implicit and explicit.  }
\label{tab:app_qa_example}
\end{table*}

\begin{figure*}[!htb]
\centering
    
    \includegraphics[width=.9\textwidth]{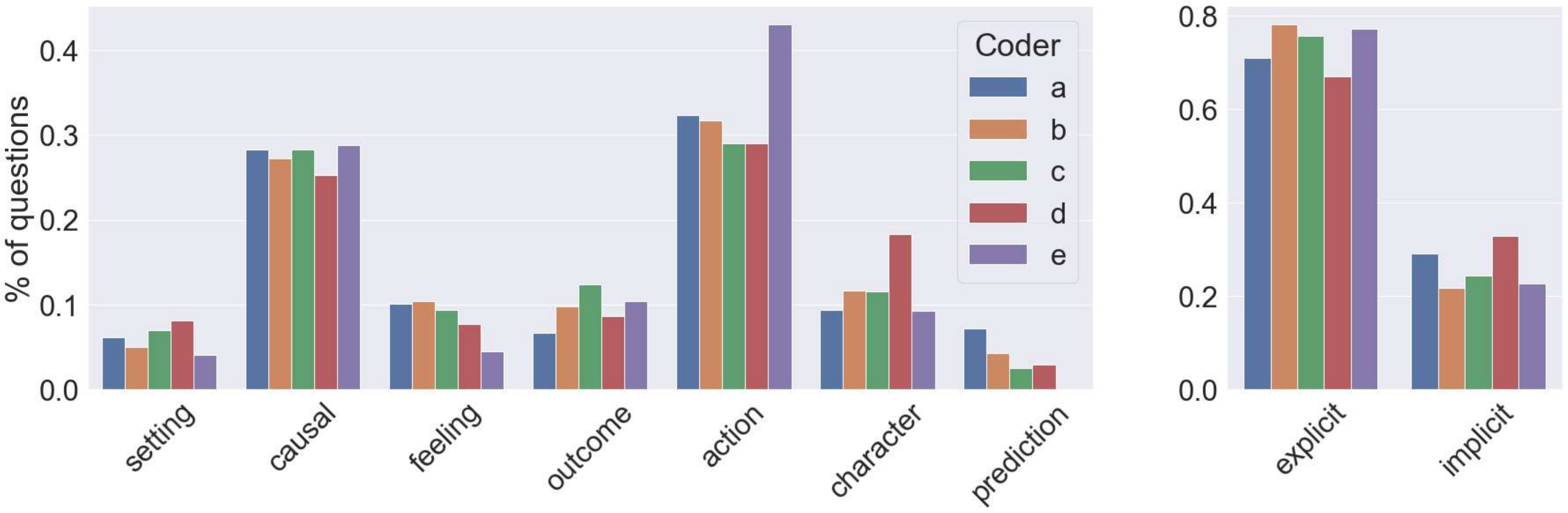}
    \caption{Percent of each question type by coder.}
    \label{fig:proportion_question}
    
\end{figure*}

Table~\ref{tab:qg_example_appeldix} shows two QG examples that have an input of event-related ground-truth answers. We may notice that BART fine-tuned on NarrativeQA is able to generate questions that seem to be in a correct format but suffer from fact error, while BART fine-tuned on \datasetname is able to generate questions that are very similar to ground-truth questions and are semantically correct. Since the crowd workers only read the abstracts to create QA-pairs in NarrativeQA, in comparison, we ask our coders to read the complete story. This may lead to an issue with models fine-tuned on NarrativeQA where the evidence of the answer in the original text content is not detailed and obvious enough for QA-pairs in NarrativeQA so that the QG model fine-tuned on NarrativeQA is not ad good as models fine-tuned on \datasetname in locating evidence.

\section{Example questions by category in \datasetname}
\label{appendix:example_qs}

Table~\ref{tab:app_qa_example} shows example QA-pairs for different annotations in \datasetname dataset. There is one example QA-pair for each narrative element as well as for implicit and explicit.

\section{Fine-tuning Parameters}
\label{appendix:parameters}
For the QA task, we keep the following fine-tuning parameters consistent over different datasets: \textit{learning rate} = $5e^{-6}$; \textit{batch size} = $1$; \textit{epoch} = $1$. \\
For the QG task, we keep the following fine-tuning parameters consistent over different datasets: \textit{learning rate} = $5e^{-6}$; \textit{batch size} = $1$; \textit{epoch} = $3$.

\label{app:att_plot}

\section{Proportion of Each Question Type}

Figure~\ref{fig:proportion_question} shows the percentage of each question type by coder.

\end{document}